\documentclass{article}
\usepackage{booktabs}
\usepackage{multirow}
\usepackage{caption}  

\usepackage[preprint]{neurips_2024}

\usepackage[utf8]{inputenc} 
\usepackage[T1]{fontenc}    
\usepackage{hyperref}       
\usepackage{url}            
\usepackage{booktabs}       
\usepackage{amsfonts}       
\usepackage{nicefrac}       
\usepackage{microtype}      
\usepackage{xcolor}         
\setcitestyle{numbers,square}

\usepackage{latexsym}
\usepackage{bm}
\usepackage{graphicx}
\usepackage{wrapfig}
\usepackage{natbib}
\usepackage{amssymb}
\usepackage{amsmath}
\usepackage{amsthm}
\usepackage{mathrsfs}
\usepackage{makecell}
\usepackage{colortbl}

\definecolor{GrassGreen}{RGB}{146,208,80}
\definecolor{SkyBlue}{RGB}{4,175,252}
\definecolor{FluoGreen}{RGB}{24,229,25}
\definecolor{Fig1aGreen}{RGB}{106, 234, 102}
\definecolor{Fig1aBlue}{RGB}{0, 176, 240}
\definecolor{EncoderBlue}{RGB}{166, 203, 220}
\definecolor{DecoderGreen}{RGB}{131, 186, 158}
\definecolor{Pending}{RGB}{255, 0, 0}
\definecolor{red}{RGB}{255, 0, 0}
\definecolor{green}{RGB}{0, 255, 0}
\definecolor{blue}{RGB}{0, 0, 255}
\definecolor{AblationRed}{RGB}{251, 221, 238}
\definecolor{AblationRedWord}{RGB}{231, 146, 238}
\definecolor{AblationGreen}{RGB}{193, 220, 206}
\definecolor{AblationGreenWord}{RGB}{160, 191, 118}

\definecolor{TableColor}{rgb}{0.835, 0.894, 0.968}
\usepackage{lipsum}		
\usepackage{doi}
\usepackage{bbding}

\title{PersonViT: Large-scale Self-supervised Vision Transformer for Person Re-Identification}

\author{%
  Bin~Hu \\
  Institute of Artificial Intelligence, Huazhong University of Science and Technolog\\
  Wuhan, China \\
  \texttt{hubin@hust.edu.cn} \\
  \And
  Xinggang~Wang \\
  School of EIC, Huazhong University of Science and Technolog \\
  Wuhan, China \\
  \texttt{xgwang@hust.edu.cn} \\
  \AND
  Wenyu~Liu\thanks{Corresponding Author.}\\
  School of EIC, Huazhong University of Science and Technolog \\
  Wuhan, China \\
  \texttt{liuwy@hust.edu.cn} \\
}

\begin{document}

\maketitle


\begin{abstract}
  Person Re-Identification (ReID) aims to retrieve relevant individuals in non-overlapping camera images and has a wide range of applications in the field of public safety. In recent years, with the development of Vision Transformer (ViT) and self-supervised learning techniques, the performance of person ReID based on self-supervised pre-training has been greatly improved. Person ReID requires extracting highly discriminative local fine-grained features of the human body, while traditional ViT is good at extracting context-related global features, making it difficult to focus on local human body features. To this end, this article introduces the recently emerged Masked Image Modeling (MIM) self-supervised learning method into person ReID, and effectively extracts high-quality global and local features through large-scale unsupervised pre-training by combining masked image modeling and discriminative contrastive learning, and then conducts supervised fine-tuning training in the person ReID task. This person feature extraction method based on ViT with masked image modeling (PersonViT) has the good characteristics of unsupervised, scalable, and strong generalization capabilities, overcoming the problem of difficult annotation in supervised person ReID, and achieves state-of-the-art results on publicly available benchmark datasets, including MSMT17, Market1501, DukeMTMC-reID, and Occluded-Duke. The code and pre-trained models of the PersonViT method are released at \url{https://github.com/hustvl/PersonViT} to promote further research in the person ReID field.
\end{abstract}

\section{Introduction}
\begin{figure}
\centering
\includegraphics[width=0.8\linewidth]{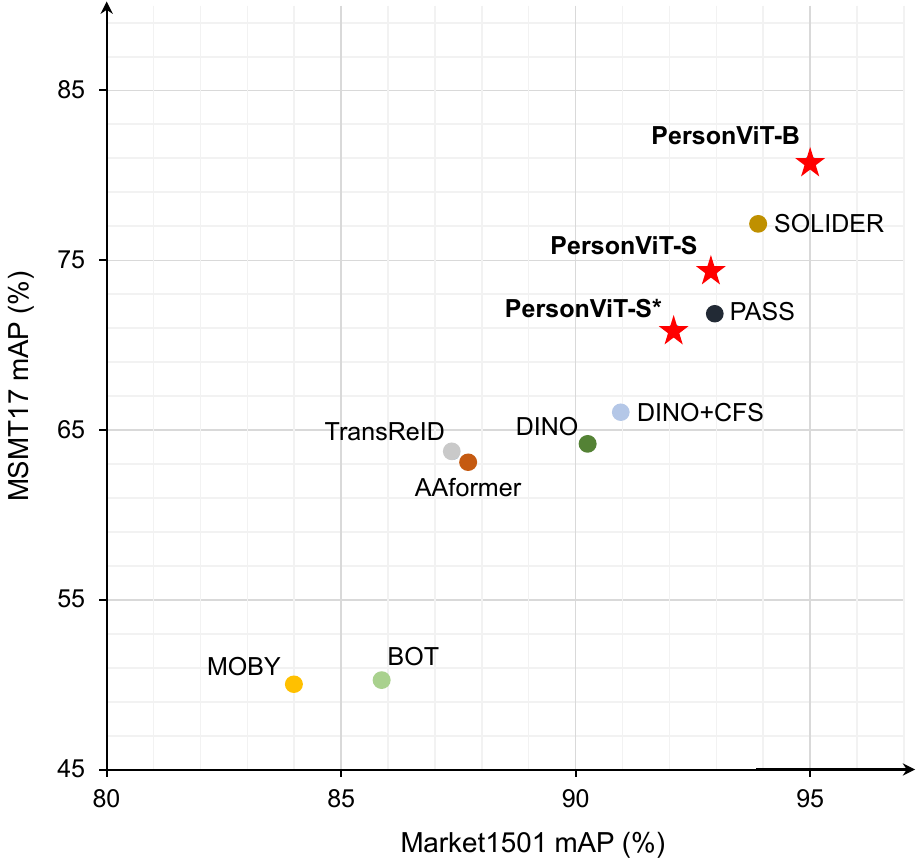}
\caption{\label{fig:compare}Person ReID performance on both MSMT17 and Market1501. The proposed PersonViT method obtains SOTA results and significantly outperforms previous methods.}
\end{figure}
    
Person re-identification (ReID) aims to learn visual features from human images that can distinguish different individual identities. It is an important and challenging computer vision problem that requires overcoming severe occlusion, appearance changes, shape changes, and viewpoint changes. Person ReID technology enables effective cross-camera retrieval of people in contactless and uncooperative scenarios and is widely used in public safety, video surveillance, and other fields, offering significant application value.  

There have been many studies related to person ReID, most of them involve metric learning (for example, Triplet Loss \cite{triplet_loss}) on features extracted by backbone networks (such as ResNet \cite{ResNet} or ViT \cite{ViT}). Constrained by the small size of labeled ReID datasets, most of the backbone networks in the current methods are pre-trained on ImageNet \cite{imagenet} and then fine-tuned on ReID datasets to achieve better performance. However, there is a significant difference between ImageNet and person ReID datasets, as ImageNet contains images of a thousand categories while ReID has only one person category. Therefore, models pre-trained on ImageNet can better extract category-level features but struggle to effectively extract fine-grained individual features, making them less suitable for the person ReID problem \cite{zhu2022part}.

To bridge the gap between pre-training and fine-tuning datasets for better ReID backbones, \cite{LUPerson} collected an unlabeled person dataset called LUPerson and demonstrated for the first time that CNN-based self-supervised pre-trained models perform well in ReID tasks. Subsequently, TransReID-SSL \cite{reid_ssl} conducted experiments using ViT-based self-supervised learning algorithms on the LUPerson dataset and found that models pre-trained with the DINO \cite{DINO} algorithm were most effective for person ReID. However, these self-supervised learning algorithms all utilize the classification consistency between local and global images generated from different data augmentations of the same image to perform contrastive self-supervised learning. They can capture the inter-class differences in multi-class datasets such as ImageNet relatively well, but lose the fine-grained differences in single-class ReID datasets. For example, local images of different human bodies may be very similar and can belong to the same class, but DINO's self-supervised contrast will separate them into different classes. Similarly, local images of the same person at different angles should belong to the same class, but DINO's contrast will separate them. Therefore, the PASS method \cite{zhu2022part} proposed a self-supervised pre-training algorithm designed specifically for person ReID tasks. Based on DINO, it divides person images into several (L) blocks according to the human body structure from top to bottom. It performs contrastive self-learning between blocks and between blocks and the whole image, thus enhancing the pre-trained model's ability to express local features and achieving new state-of-the-art at the time. However, PASS has two problems: 1) like the previous block-based ReID such as PCB \cite{PCB}, it has an alignment dependency problem, i.e., the block division method is too mechanical in complex backgrounds, non-uniform alignments, and even occluded or incomplete human images, making it prone to misclassification; 2) a smaller number of blocks still struggle to fully express fine-grained local features, i.e., in the PASS method, the optimal performance experiment corresponds to a partition count of only three, which is evidently insufficient in granularity.

Recently, inspired by the monumental success of Masked Language Modeling (MLM), i.e. BERT \cite{bert}, in the Natural Language Processing (NLP) field, Masked Image Modeling (MIM) techniques such as BEiT \cite{bao2021beit}, MAE \cite{he2022mae}, and SimMIM \cite{xie2022simmim} has also made breakthroughs in image classification, detection, and segmentation tasks in natural images through self-supervised learning. MIM employs random masking of local pixels for reconstruction learning, reinforcing the learning and extraction of local fine-grained features without manual image partitioning. This effectively compensates for the limitations of the PASS algorithm, indicating that the introduction of MIM should enhance the accuracy of person ReID. 

In this paper, we incorporate a MIM self-supervised learning module based on DINO, conducting large-scale, unlabeled pre-training on the LUPerson \cite{LUPerson} dataset, and subsequently fine-tuning supervised ReID on four datasets: MSMT17 \cite{MSMT17}, Market1501 \cite{Market1501}, DukeMTMC-reID \cite{DukeMTMC-reID}, and Occluded-Duke \cite{miao2019pose}. We refer to this method of utilizing masked image modeling and DINO contrastive learning for large-scale unsupervised pre-training based on vanilla ViT and fine-tuning on downstream datasets as the PersonViT method. Experimental results reveal that PersonViT achieves state-of-the-art results, particularly excelling on the challenging dataset Occluded-Duke. Furthermore, visualization analysis of the pre-trained models on MSMT17 discloses that PersonViT can automatically discover key human body parts, clothing patterns, and the correspondence between local body parts without annotations.

The innovations of this paper encompass the following aspects: 1) To reinforce the algorithm's robustness to human image alignment and better extract the local fine-grained features of the human body, we pioneer the utilization of masked image modeling techniques in unsupervised feature learning for person ReID. This effectively addresses the challenge of acquiring local fine-grained visual features in human images with substantial occlusions and misalignments. 2) We propose a highly efficient, large-scale self-supervised human feature learning method based on the vanilla ViT, referred to as PersonViT. 3) Our approach achieves state-of-the-art results on several mainstream person ReID datasets.

The main significance of this research is that it achieves industry-leading ReID results through efficient and accurate self-supervised feature learning, greatly addressing the dilemma of insufficient labeled training data in person ReID. The proposed method is highly scalable and provides strong technical support for large-scale practical applications of ReID technology. Experimental results show that our method achieves industry-leading accuracy, significantly surpassing previous approaches.

\section{Related Work}
\subsection{Self-Supervised Learning}
\paragraph{Contrastive Learning} Self-supervised learning (SSL) methods aim to learn discriminative features from large-scale unlabeled data \cite{jing2020self}. In recent years, contrastive learning methods have flourished in the field of computer vision \cite{MoCo_v1,MoCo_v2,MoCo_v3,simCLR,BYOL,MoBY,DINO}, significantly narrowing the gap with supervised pretraining. MOCO \cite{MoCo_v1} first proposed momentum contrast, in which a pair of enhancements of a sample is regarded as positive, while other samples and their enhancements as negative, thereby conducting unlabeled contrastive learning training. This approach has been followed by a series of improvements such as MOCOV2 \cite{MoCo_v2}, MOCOV3 \cite{MoCo_v3}, and SimCLR \cite{simCLR}. BYOL \cite{BYOL} proposed a new contrastive learning paradigm that uses two network models to predict the representations of different enhanced images of the same sample, eliminating the dependence on large batches of negative samples. DINO \cite{DINO} is an improved version of BYOL. It introduces centering and sharpening operations into the process of momentum smoothing update of the target model parameters, effectively preventing model collapse and improving the algorithm's stability. In addition, DINO incorporates extensive data enhancements, especially multiple local cropping enhancements, which enhance the model's ability to learn local features through large-scale contrastive learning between local and global images.

\paragraph{DINO Framework}
DINO \cite{DINO} serves as the inaugural self-distillation learning framework. Ahead of delving into DINO, let's briefly touch upon the foundational principles behind Transformer. Viewed as an encoder, the Transformer translates images $I \in \mathbb{R} ^{h \times w \times c}$, where $h \times w$ corresponds to the image resolution and $c$ to the number of image channels, typically 3, denoting RGB, into target feature vectors. Drawing parallels with tokenization procedures in NLP, the image is bucketed into continuous entities. Transformers initiate the process through block projection tokenization of the images. If each $p \times p$ pixel-sized block is partitioned into $n=hw/p^2$ image patches, every segment can be depicted as a token, bearing similarity to individual words in NLP. Coupled with a learnable $cls\_token$, these tokens can represent image tokenization results as $X$ as indicated in Eq.~\eqref{eq:tokens}. Here, the ";" symbol symbolizes the connection between stacked tokens. Following encoding via the ViT network, a similar representation can be achieved for feature vectors as per Eq.~\eqref{eq:vit_features}. Contrasting with the conventional ViT backbone network, the DINO encoder integrates a head module that channels the ViT output $z^{[cls]}$ through a multi-layer perception network (MLP) to map onto the targeted vector space $y^{[cls]}$ as illustrated in Eq.~\eqref{eq:dino_header}. To calculate the ultimate loss function, denoted as $L_{dino}$ for differentiation from subsequent loss functions, DINO utilizes two isomorphic encoder networks - the student and teacher network, their respective outputs denoted as $Y^{[S]}$ and $Y^{[T]}$ as indicated in Eq.~\eqref{eq:dino_loss}. The norm of gradient backpropagation updates the student network parameters, whereas the teacher network parameters evolve in tandem with the Exponential Moving Average (EMA) of the student network parameters, calculated as per $\theta_t = \lambda\theta_t + (1-\lambda)\theta_s$. In an interesting twist, the DINO teacher network inputs two globally amplified data views while the student network calls upon corresponding global views as well as multiple local views. Consequently, $L_{dino}$ encapsulates both global and local-global contrast calculations, thereby enhancing the discernment of local features.

\begin{align}
\begin{split}
    X &= (x^{[cls]};\ x^{[patchs]})\\
      &= (x^{[cls]};\ x_1; \dots;\ x_n) \in \mathbb{R}^{(n+1) \times d}
\end{split}
\label{eq:tokens}
\end{align}

\begin{align}
\begin{split}
    Z &= (z^{[cls]};\ z^{[patchs]})\\
      &= (z^{[cls]};\ z_1; \dots;\ z_n) \in \mathbb{R}^{(n+1) \times d}
\end{split}
\label{eq:vit_features}
\end{align}

\begin{equation}
    Y = \text{MLP}(z^{[cls]}) = y^{[cls]} \in \mathbb{R}^{1 \times d_{output}}
\label{eq:dino_header}
\end{equation}
\begin{equation}
    L_{dino} = -\text{Softmax}(Y^{[T]})\log(\text{Softmax}(Y^{[S]}))
\label{eq:dino_loss}
\end{equation}

\paragraph{Masked Image Modeling} Starting from 2018, the field of NLP has witnessed notable success in Masked Language Modeling (MLM), represented by models such as BERT \cite{bert} and GPT \cite{gpt}. In recent years, several masked modeling attempts have also emerged in the field of vision, with representative works including \cite{li2021mst, xie2022simmim, he2022mae, zhou2021ibot, assran2022msn}. MST\cite{li2021mst}, the first model to introduce masked image modeling, enhances the performance of DINO by adding masked predictive training to the DINO framework. Following closely is BEiT \cite{bao2021beit}, which first uses a discrete Variational AutoEncoder (dVAE) to discretize image blocks and map them to corresponding vision tokens. It then masks the obstructed vision tokens to achieve prominent learning results; therefore, BEiT also operates as a two-stage training algorithm.
MAE \cite{he2022mae} is an end-to-end MIM autoencoder. It employs an encoder to code only visible image blocks and utilizes a relatively lightweight decoder to restore masked pixels. Interestingly, it has been shown to produce outstanding self-supervised learning results even when a majority of the input image (e.g., 75\%) is masked, thereby significantly accelerating the pre-training process. As MAE's self-supervised learning is to restore pixels, both its encoder and decoder learn certain feature expressions. Moreover, MAE restores the most primitive low-level pixel features, so the degree of feature abstraction learned by the encoder is not controlled. The relatively low accuracy of the k-NN and linear probing benchmarks in MAE's ImageNet experiment results signifies that the encoder does not sufficiently extract distinguishing feature abstracts.

\subsection{Part-based person ReID}
Traditionally, feature expression learning predominantly employed an IDE model \cite{zheng2017person}. This approach involved global feature extraction followed by multi-classification training, where each individual was treated as a separate class. Gradually, the trend evolved to use ResNet50 \cite{ResNet} to extract global feature vectors, which were then subjected to metric learning, such as Triplet Loss \cite{triplet_loss}.
However, extracting detailed local features remained a challenge for global feature representation learning. As a result, two primary forms of optimization emerged to address this. The first involved multi-scale fusion representation learning as presented by documents such as \cite{qian2017multi}. The second included the integration of attention mechanisms to strengthen local representation learning, such as harmonizing spatial and channel attention \cite{li2018harmonious}, and attenuating background in representation learning using segmentation attention mechanism \cite{song2018mask}.
To better capture local, fine-grained features of human images and improve the robustness of algorithms in handling alignment issues and obstructions, local feature-based person ReID methods were subsequently proposed. Early proposed local representation learning methods, such as those found in \cite{PAR}, \cite{part_bilinear}, and PCB \cite{PCB}, partitioned human images into multiple horizontal stripes to extract local features.
MGN \cite{MGN} enhanced the robustness using different granularity stripes and overlaps between these stripes. TransReID \cite{transreid}, being the first Transformer-based person ReID algorithm, also involves local representation learning by rearranging and regrouping Transformer's input local images. These methodologies highlight that the nuanced features extracted through local representation learning significantly bolster the precision of person ReID.

\subsection{Self-supervised person ReID} Although research related to person ReID in self-supervised pre-training surfaced relatively late and is therefore scarcer, it showcases tremendous potential due to the avoidance of high-cost annotations in the pre-training phase. TransReID-SSL \cite{reid_ssl} was the first to compare the performance of the then mainstream self-supervised learning algorithms on the problem of ReID. In the comparison, DINO \cite{DINO} demonstrated significant advantages over other algorithms. The researchers also studied the effect of pre-training data size on final accuracy and proposed an effective method for screening pre-training data. PASS \cite{zhu2022part}, the first self-supervised pre-training algorithm geared towards the ReID task, segments human images. Building on the foundation of DINO, PASS introduced local-to-local and local-to-global contrast learning. It differentiates representations by using separate $cls\_token$s for different locals, and in addition to the global contrast classification feature space[CLS], also introduces additional segment classification feature space[PART]s, thereby enhancing the detailed representation capability of segments and improving local feature extraction ability of the pre-training model. The method set industry-leading benchmarks at the time in both supervised person ReID and cross-domain person ReID, thereby validating its effectiveness.

\subsection{Summary of Research on person ReID}

Reflecting on the evolution of person ReID over the past few years, we can summarize three major trends.

Firstly, the feature extraction backbone network has evolved from traditional convolutional neural networks to Transformers (ViT), primarily due to the advanced capabilities of ViT to express global contextual features and the significant developments in ViT-based self-supervised learning.

Secondly, the initial pre-training approaches based on ImageNet classification have moved towards self-supervised models using extensive human data. This shift is mainly because ReID challenges involve distinguishing minor differences between different individuals, unlike ImageNet classification tasks with varied and broad class differences. As the test individuals are typically not present in the training subsets, the generalization capability forms a pivotal component driving recognition accuracy. Nevertheless, the scarcity of public supervised training sets, resulting from challenges in data acquisition and labeling, falls short in fulfilling the requirements for algorithmic generalization capability.

Lastly, given the nuanced distinctions between different individuals—a much smaller scope compared to class variations in ImageNet—effective extraction of local fine-grained features proves crucial in enhancing ReID accuracy.

The latest methodology, PASS \cite{zhu2022part}, successfully aligns with these three trends and achieved industry-leading scores at the time of its inception. However, it highlights two critical limitations. Firstly, its reliance on manually designed segment partitioning restricts its robustness to the alignment of human images. Secondly, the finite number of partitions makes it challenging to adequately express local fine-grained features, thereby constraining the capability of the pre-training model to extract such features.

\section{Method}
Addressing the limitations identified in the PASS algorithm, a potential solution might be to introduce a classification feature space to each image patch input in the ViT model, akin to the [CLS] feature space, in the context of self-supervised pre-training. However, applying traditional data augmentation-based contrastive learning methods might pose challenges. Given the small size of each patch, the process of generating different data augmentations could easily lead to misplacements within local image patches, limiting comparability and not achieving the desired self-supervised effects. This is evidenced by the experimental results from the PASS method, which demonstrated that increasing the number of local segments or reducing the size of partitions does not consequently improve accuracy.

Inspired by the concept of masked image modeling, a potential solution might be to feed the same image into the system twice. One input would be the complete image, and the other would be the image with some patches masked. The corresponding output feature vectors from both inputs, denoted [PATCH] and represented as $\tilde{y}^{[patchs]}$ in Eq.~\eqref{eq:mim_header}, would form a basis for comparison. This approach allows for feature representation at the most granular level, corresponding to the size of the patch, and could improve the pre-training model's ability to extract fine-grained local features.

\begin{figure*}
\centering
\includegraphics[width=.8\linewidth]{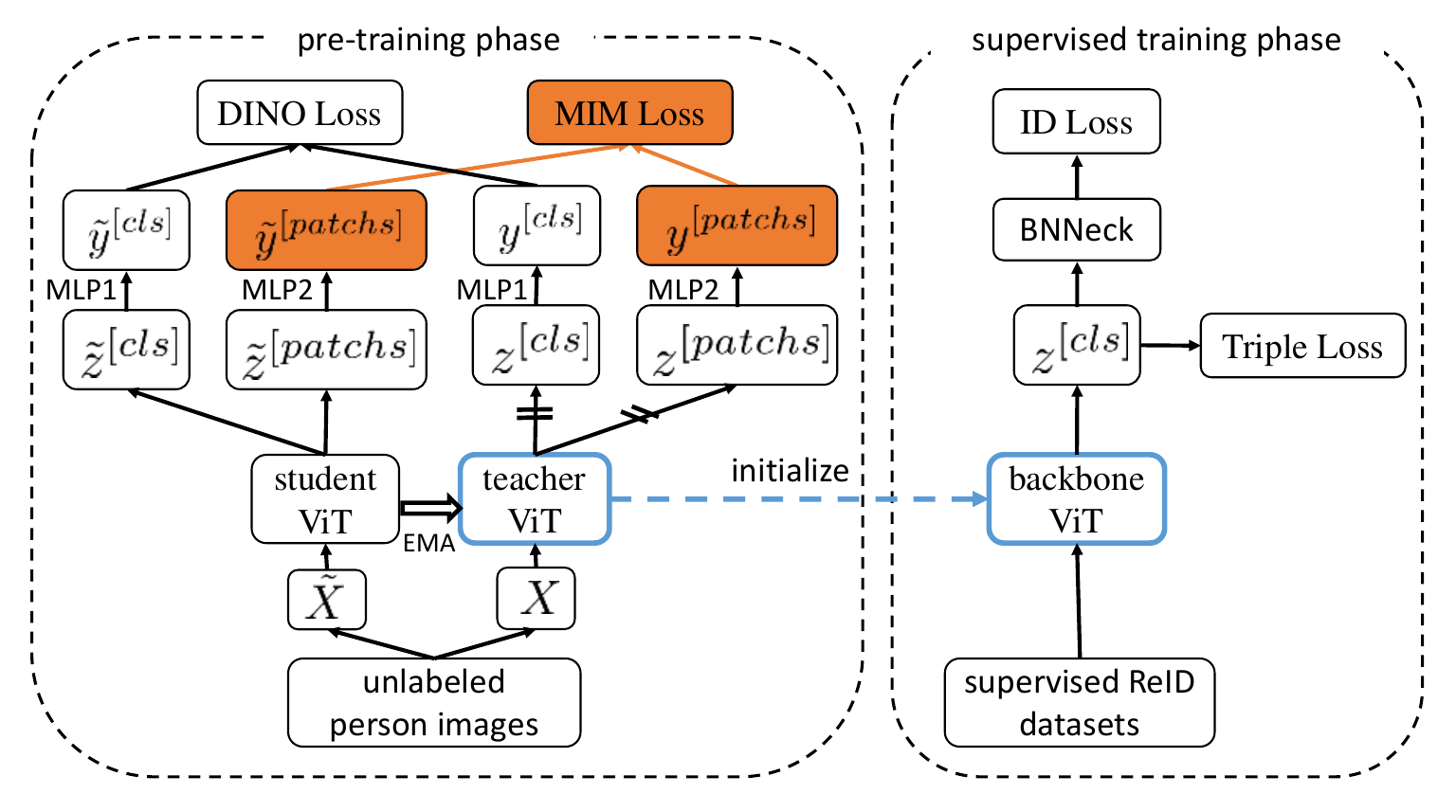}
\caption{\label{fig:main_arch}Overview of PersonViT framework.}
\end{figure*}

The overall algorithm framework proposed in this paper, depicted in Fig.~\ref{fig:main_arch}, incorporates a contrast loss module at the patch granularity level (known as the MIM Loss module) into the DINO pre-training algorithm. Pre-training is executed with ViT-S and ViT-B as the backbone networks utilizing the largest publicly available dataset, LUPerson \cite{LUPerson}. This is followed by supervised fine-tuning training on four mainstream public person ReID datasets: MSMT17 \cite{MSMT17}, Market1501 \cite{Market1501}, DukeMTMC-reID \cite{DukeMTMC-reID}, and Occluded-Duke \cite{miao2019pose}. The procedures for both pre-training and fine-tuning are detailed below.

\subsection{Incorporation of Self-Supervised Pre-training MIM Loss Function}
Inspired by the masked image modeling paradigm of BEiT \cite{bao2021beit}, our methodology encompasses the random block-wise masking of images, akin to the introduction of a learnable token variable $x^{[mask]}$, analogous to $x^{[cls]}$. Consequently, the masked image may be delineated as $\tilde{X} =(x^{[cls]}; \tilde{x}_1; \dots; \tilde{x}_n)$ with $\tilde{x}_i$ detailed in Eq.~\eqref{eq:mask_x}. 
\begin{align}
\begin{split}
    \tilde{X} &=(x^{[cls]}; \tilde{x}_1; \dots; \tilde{x}_n), \\
    \tilde{x}_i &= (1 - m_i)x_i + m_ix^{[mask]}, i\in 1, \dots, n
\end{split}
\label{eq:mask_x}
\end{align}
In this equation, $m_i \in 0, 1$ signifies the random image block mask, where 1 corresponds to masked and 0 to unmasked. The masked image $\tilde{X}$, when processed through the ViT encoder, yields $\tilde{Z}$, as depicted in Eq.~\eqref{eq:vit_features}. 
\begin{align}
\begin{split}
    \tilde{Y} &= (\text{MLP1}(\tilde{z}^{[cls]});\   \text{MLP2}(\tilde{z}^{[patchs]}))\\
              &= (\tilde{y}^{[cls]};\ \tilde{y}^{[patchs]})
    \in \mathbb{R}^{(1+n) \times d_{output}}    
\end{split}
\label{eq:mim_header}
\end{align}

\begin{equation}
\begin{split}
  L_{mim}  &= \sum_{i=1}^n m_i \cdot P(y_i^{[patchs][T]}) \log(P(\tilde{y}_i^{[patchs][S]})), \\
   P(x) &= \text{Softmax}(x)
\end{split}
\label{eq:loss_mim}
\end{equation}

\begin{equation}
  L = \lambda_1 L_{dino} + \lambda_2 L_{mim}
\label{eq:loss_total}
\end{equation}
Strikingly different from DINO, here we also implement an MLP network transformation for $\tilde{z}^{[patchs]}$, thereby projecting it onto a $d_{output}$-dimensional vector space to produce $\tilde{y}^{[patchs]}$, prior to executing the reconstruction loss calculation. The target vector $\tilde{Y} =(\tilde{y}^{[cls]}; \tilde{y}_1; \dots; \tilde{y}_n)$ can then be articulated as shown in Eq.~\eqref{eq:mim_header}. To benchmark the reconstructed vector of the masked image blocks, masking is solely applied to the dual global view inputs of the student network, with the teacher network's input comprising an intact global view for reference. The specific masked reconstruction loss function (abbreviated as $L_{mim}$ is explicated in Eq.~\eqref{eq:loss_mim}. This, in conjunction with the DINO loss function $L_{dino}$ as evinced in Eq.~\eqref{eq:dino_loss}, forms the final pretraining loss function $L$ as outlined in Eq.~\eqref{eq:loss_total}. Here, $L$ is conceived as the weighted sum of $L_{dino}$ and $L_{mim}$, wherein $\lambda_1=\lambda_2=1$ by default.

\subsection{Supervised Fine-tuning}
In this stage, the pre-training model produced from prior self-supervised learning undergoes fine-tuning for the specific task of person ReID, generating the final model for this task. Subsequently, we conduct tests to evaluate the ReID accuracy of the model. To ensure unbiased comparison of the efficacy of the self-supervised pre-training model, we continue utilising the BOT \cite{BOT} framework during this stage. This approach aligns with the methodology employed in the baseline of TransReID-SSL, which consists of using a standard ViT network (i.e., ViT-S/16 and ViT-B/16) for the primary network, directly implementing $z^{[cls]}$ for feature aggregation, adopting Triplet Loss for metric learning, choosing Cross-entropy Loss for ID Loss, and inserting a BNNeck module between metric learning and ID Loss. For the primary standard ViT network, we initiate fine-tuning for person ReID training by using the pre-training model from the teacher network in the self-supervised learning stage as the starting reference point.

\section{Experiments}
\subsection{Datasets}
The main dataset used for self-supervised pre-training is LUPerson \cite{LUPerson}, which includes 4.18 million unlabeled human images. This dataset is four times the size of ImageNet, therefore, the computational resources required for pre-training are larger, and the same computational resources will result in four times the training time of the ImageNet dataset. To thoroughly verify the pre-training model's ability to extract finer-grained local features, this research conducts supervised training on the four mainstream person ReID datasets, MSMT17 \cite{MSMT17}, Market1501 \cite{Market1501}, DukeMTMC-reID \cite{DukeMTMC-reID}, and Occluded-Duke \cite{miao2019pose}, observing boosts in two key indicators: mAP (mean average precision) and Rank-1. The detailed information of the four datasets is shown in Table~\ref{tab:datasets}. Occluded-Duke, which is generated based on DukeMTMC-ReID, enhances the difficulty of ReID under occlusion situations.

\begin{table}[h!t]
\centering
\caption{\label{tab:datasets}Statistics of some commonly used datasets for person ReID.}
\begin{tabular}{l|l|l|l|l}
\hline
Dataset        & Time & \#ID  & \#image    & \#cam \\ \hline
Maket-1501 & 2015 & 1501 & 32668  & 6    \\ \hline
DukeMTMC   & 2017 & 1404 & 36411  & 8    \\ \hline
MSMT17     & 2018 & 4101 & 126441 & 15   \\ \hline
Occluded-Duke   & 2019 & 1404 & 36411  & 8    \\ \hline
\end{tabular}
\end{table}

\subsection{Implementation Details}

\subsubsection{Self-Supervised Pre-Training Stage} 
Due to the large size of the LUPerson dataset and to reduce the experimental timeframe, we implemented large-batch training using 8$\times$8$\times$A100 GPUs to expedite our experiments. Nonetheless, to demonstrate the effectiveness of small batch training, we also carried out foundational experimentation using 4$\times$RTX3090 GPUs. To minimize the computational load during the pre-training phase, the number of training cycles (epochs) was set to 300. Similar to DINO and PASS, the teacher network accepted input with image dimensions of 256$\times$128, whereas the student network was designed to process global views with dimensions of 256$\times$128 and only 6 local views with dimensions of 96$\times$64. Considering the extended time taken for pre-training, our experiments only trained two basic network models - ViT-S/16 and ViT-B/16. For the small batch experiment, the pre-training learning rate uniformly adopts $lr=0.0005 * batch\_size/256$. For large-batch experiments, despite a significant reduction in experimental duration, the training process was rather unstable. Consequently, we adjusted the learning rate based on the specific batch size, making sure not to exceed 0.002, until stable training convergence was achieved. The specific parameters and training logs will be publicly released along with the code.

\subsubsection{Supervised Training Stage}
Based on the experimental setup of TransReID-SSL \cite{reid_ssl}, the supervised fine-tuning was performed without adding any other optimization items, only using traditional ViT-S/16 and ViT-B/16 Transformer networks as the backbone for supervised training. The supervised training uniformly adopted stochastic gradient descent as the learning algorithm, and the learning rate was set as $lr=0.0004 * batch\_size / 64$. The batch size was composed of $4 * 16=64$, which means 16 different individuals per batch, with 4 images for each person. Following the same strategy as PASS \cite{zhu2022part}, the first 20 cycles served as a warm-up phase, and the $\alpha$ parameter for Triplet Loss was set at 0.3.

\subsection{Experimental Results}
The comparison of different algorithms' experimental results is shown in Table~\ref{tab:main_results} (the light blue background represents the highest accuracy). To make a fair comparison about the role of the pre-training parameters of the backbone network, TransReID- represents the ReID accuracy without the SIE and JPM modules from TransReID \cite{transreid}. Furthermore, the last two rows represent the accuracy of the models pre-trained with 64xA100 large batch size. The row indicated with a * represents the accuracy of models pre-trained with a small batch size on 4xRTX3090. AAformer \cite{AAformer} and TransReID- are models pre-trained for classification on the ImageNet-21K dataset, which is larger than the LUPerson dataset and has classification labels. As seen, the ReID accuracy of the self-supervised pre-training models based on LUPerson far exceeds those based on ImageNet pre-training models.

Comparing with DINO+CFS \cite{reid_ssl} and PASS \cite{zhu2022part}, the LUPerson pre-training models based on PersonViT proposed in this paper, and particularly the large batch size pre-training models, achieve significant improvements in accuracy, greatly surpassing the model pre-trained with the PART algorithm based on image partitioning self-supervised learning.

\begin{table}
\center
\caption{\label{tab:main_results}Comparison with the state-of-the-art methods.}
\resizebox{0.98\textwidth}{!}{
\begin{tabular}{l|l|ll|ll|ll|ll} 
\toprule
\multicolumn{1}{c}{\multirow{2}{*}{Methods}} & \multirow{2}{*}{Backbone} & \multicolumn{2}{l}{MSMT17} & \multicolumn{2}{l}{Market1501} & \multicolumn{2}{l}{DukeMSMT} & \multicolumn{2}{l}{Occluded-Duke} \\
\multicolumn{1}{c}{}                    &                       & mAP        & Rank-1       & mAP          & Rank-1         & mAP          & Rank-1        & mAP            & Rank-1           \\ 
\midrule
BOT \cite{BOT} \tiny{CVPRW2019}                               & R50          & 50.2       & 74.1         & 85.9         & 94.5           &        -      &      -         &      -          &            -      \\
AAformer \cite{AAformer} \tiny{Arxiv}                               & ViT-B/16$\uparrow$384          & 63.2       & 83.6         & 87.7         & 95.4           &        -      &      -         &      -          &            -      \\
TransReID$^-$ \cite{transreid} \tiny{ICCV2021}                              & ViT-B/16              & 63.6       & 82.5         & 87.4         & 94.6           &     -         &         -      &         -       &     -             \\ \hline
MOBY \cite{MoBY} \tiny{Arxiv}                                   & ViT-S/16              & 50.0       & 73.2         & 84.0         & 92.9           &     -         &         -      &  -              &      -            \\
DINO \cite{DINO} \tiny{ICCV2021}                                   & ViT-S/16              & 64.2       & 83.4         & 90.3         & 95.4           &     -         &         -      &  -              &      -            \\
DINO+CFS \cite{reid_ssl} \tiny{Arxiv}                               & ViT-S/16              & 66.1       & 84.6         & 91.0         & 96.0           &    -          &      -         &   -             & -                 \\
PASS \cite{zhu2022part} \tiny{ECCV2022}                                    & ViT-S/16              & 69.1       & 86.5         & 92.2         & 96.3           & 82.5         & 90.7          & 60.2           & 70.4             \\
PASS \cite{zhu2022part} \tiny{ECCV2022}                                  & ViT-B/16              & 71.8       & 88.2         & 93.0         & 96.8           & 84.7         & 92.5          & 64.3           & 74.0             \\
SOLIDER \cite{chen2023beyond} \tiny{CVPR2023}                                  & Swin-B              & 77.1       & 90.7         & 93.9         & 96.9           & -         & -          & -           & -             \\
PersonMAE \cite{hu2024personmae} \tiny{TMM2024}                                  & ViT-S/16              & 75.2       & 89.1         & 92.5         & 96.7           & -         & -          & 65.2           & 72.0             \\ 
PersonMAE \cite{hu2024personmae} \tiny{TMM2024}                                  & ViT-B/16              & 79.8       & 91.4         & 93.6         & 97.1           & -         & -          & 69.5           & 76             \\ 

\hline
PersonViT(ours)*                              & ViT-S/16              & 70.9       & 87.3         & 92.1         & 96.4           & 83.6         & 91.6          & 61.5           & 70.8             \\

PersonViT(ours)                             & ViT-S/16              & 74.3  &  89.2 & 92.9    & 96.8    & 84.7    & 91.9   & 65.2      & 73.2             \\

PersonViT(ours)                             & ViT-B/16              & 80.8  & 92.0  & 95.0    & 97.6    & 88.1    & 93.8   & 72.2      & 79.8  \\ 
\bottomrule           
\end{tabular}
}
\label{tab1}
\end{table}

\subsubsection{Hyper-parameter Ablation Experiment}
The DINO study \cite{DINO} demonstrated a notable influence of multi-crop techniques on performance. Considering the distinct characteristics of pedestrian images, it becomes relevant to investigate the implications of various hyper-parameters in global and local cropping. The person ReID pre-training dataset LUPerson varies from ImageNet in two primary aspects: 1) human images often possess lower resolution, leading to less pronounced local granular details; 2) human body crops typically exhibit a rectangular shape, therefore during pre-training, images are scaled down to 256x128 for network input. This contrasts with the ImageNet's practice of using square images of 224x224 pixels as input. Acknowledging these differences, the ablation experiment primarily focuses on studying the influence of size distribution and aspect ratio variance between global and local crops. To save time in experimentation, we opted for the top 3\% of the LUPerson sub-dataset, as ranked by the CFS in TransReID-SSL \cite{reid_ssl}, as our pre-training dataset. Further, we used the ImageNet pre-training model as initial parameters. The experiment results are presented in Table~\ref{tab:params}. Here, "crop rate range" denotes the size range of local crops relative to global ones. For example, the first entry "0.1~0.6, 0.6~1.0" implies that the ratio of the local crops input for the student network within a random interval of [0.1, 0.6], while the ratio of  global crops within the random interval [0.6, 1.0].

The findings from the ablation experiment suggest that overlapping between local and global crops during self-distillation fosters characteristics more fitting for person ReID. Moreover, maintaining the aspect ratio (height: weight) at 2:1 during random jittering (the default parameter is 1:1) proves to be more beneficial. This parameter adjustment significantly enhanced the mean average precision (mAP) score by two percentage points.
\begin{table*}[h!t]
\centering
\caption{\label{tab:params}Ablation studies about the Multi-Crop Augmentation.}
\begin{tabular}{l|l|l|l|ll}
\toprule
Input size   & Crop rate range                                                                & Aspect(width/height)    & Crop size    & mAP  & Rank-1 \\ 
\midrule
192x96  & 0.1$\sim$0.6, 0.6$\sim$1.0 & 3/4$\sim$4/3 & 96x64 & 60.2 & 80.7   \\ \hline
192x96  & 0.1$\sim$0.8, 0.8$\sim$1.0 & 3/4$\sim$4/3 & 96x64 & 59.9 & 80.8   \\ \hline
192x96  & 0.1$\sim$0.8,  0.4$\sim$1.0 & 3/4$\sim$4/3 & 96x64 & 61.3 & 81.9   \\ \hline
256x128 & 0.1$\sim$0.8,  0.4$\sim$1.0 & 3/4$\sim$4/3 & 96x64 & 62.5 & 82.4   \\ \hline
\rowcolor{TableColor}
256x128 & 0.1$\sim$0.8,  0.4$\sim$1.0 & 3/8$\sim$2/3 & 96x64 & 64.7 & 83.5 \\ \hline
256x128 & 0.1$\sim$0.8,  0.4$\sim$1.0 & 3/8$\sim$2/3 & 96x48 & 64.2 & 83.2 \\ \hline
256x128 & 0.1$\sim$0.8,  0.4$\sim$1.0 & 3/8$\sim$2/3 & 128x64 & 64.4 & 83.3 \\ 
\bottomrule
\end{tabular}

\end{table*}

\subsubsection{MIM Loss Function Ablation Experiment}
To validate the role of masked image modeling, we carried out a simple ablation experiment. Specifically, under completely identical other experimental parameters, we compared the settings of $\lambda_2=1$ and $\lambda_2=0$ in Eq.~\eqref{eq:loss_total}. The experimental results are shown in Table~\ref{tab:mimab}. The results indicate that the introduction of the MIM loss function has led to a significant improvement compared to the DINO algorithm \cite{DINO}. The increase in mAP for MSMT7 reaches an impressive 6.4, far exceeding the increase yielded by the PASS approach \cite{zhu2022part} when compared to DINO (3.0). The visualization analysis of local feature clustering in Fig.~\ref{fig:vis_patch} more clearly demonstrates that our method extracts richer local fine-grained human body features, thereby validating our theoretical hypothesis.

\begin{table*}[h!t]
\centering
\caption{\label{tab:mimab}Ablation studies about the MIM Loss.}
\resizebox{0.98\textwidth}{!}{
\begin{tabular}{l|l|ll|ll|ll|ll}
\toprule
\multicolumn{1}{c}{\multirow{2}{*}{Methods}} & \multirow{2}{*}{Backbone} & \multicolumn{2}{l}{MSMT7} & \multicolumn{2}{l}{Market501} & \multicolumn{2}{l}{DukeMSMT} & \multicolumn{2}{l}{Occluded-Duke} \\
\multicolumn{1}{c}{}                    &                       & mAP        & Rank-1       & mAP          & Rank-1         & mAP          & Rank-1        & mAP            & Rank-1           \\ 
\midrule
\rowcolor{TableColor}
PersonViT                             & ViT-S/16              & 74.3  & 89.2  & 92.9    & 96.8    & 84.7    & 91.9   & 65.2      & 73.2             \\ \hline
PersonViT w/o. MIM                    & ViT-S/16              & 67.9  & 85.6  & 91.4    & 96.3    & 82.1    & 90.6   & 58.5      & 66.9 \\ 
\bottomrule
\end{tabular}
}

\end{table*}

\subsubsection{Overfitting Problem in Pre-training}
\begin{figure*}[h!t]
\centering
\includegraphics[width=0.7\linewidth]{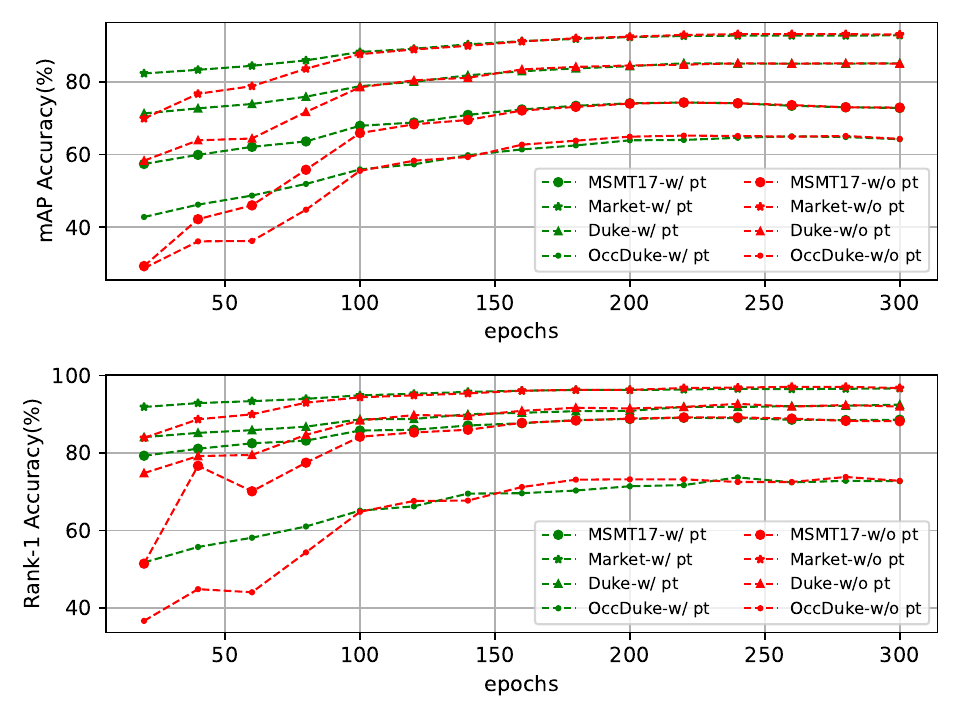}
\caption{\label{fig:overfit}Supervised accuracy of Person Re-ID on the different pre-trained epochs.}
\end{figure*}

During the experiment, we found that the optimal model accuracy is not necessarily achieved at $epoch=300$ , corresponding to the end of the pre-training process. By monitoring the metrics during the self-supervised pre-training phase, we noticed an overfitting phenomenon occurring in the epoch range [200,300]. To further investigate this overfitting issue in self-supervised learning, we undertook supervised training and tested the accuracy changes of models saved every 20 epochs under two conditions: pre-training from scratch and initializing parameters with the ImageNet self-supervised pre-training model. These results are shown in Fig.~\ref{fig:overfit}, where "w/ pt" denotes training with the full LUPerson dataset starting from ImageNet pre-training model initialization, while "w/o pt" signifies pre-training starting from zero. Two conclusions can be drawn from Figure 3: 1) Initializing with the ImageNet pre-training model gives a substantial advantage in the early stages, but this advantage is overtaken around $epoch=160$ and the pre-training from zero shows a slight edge for later epochs in the range $epoch \in [160, 300]$. The early advantage is expected as the ImageNet pre-training model already has a good discriminative capability for pedestrians. 2) For the later training phase $epoch \in [160, 300]$, both mAP and Rank-1 accuracy metrics exhibit an overfitting trend of initial increase followed by a drop. This is particularly evident in the MSMT17 dataset, where optimum performance is achieved at $epoch=200$ while in datasets optimal point occurs at $epoch=240$. Therefore, the supervised training accuracy of the model at pre-training $epoch=240$ is used for most of the experimental results presented in this paper. As the amount of training data is small (only 3\% of the total), there is no clear overfitting issue, hence the results in Table 3 were obtained at $epoch=300$.

\subsubsection{Influence of Pre-training Data Volume} To explore further the impact of pre-training data scales on the supervised training outcome, we designed experiments with different samples from the LUPerson dataset. The results are shown in Table~\ref{tab:datascale}. Here, "w/ PT." denotes whether the ImageNet pre-training model was utilized to initialize parameters. "10\%" represents randomly selecting 10\% of the samples from LUPerson, and "10\%+CFS" denotes selecting the top 10\% of samples as per the CFS ranking in TransReID-SSL \cite{reid_ssl}.

Analysis of Table~\ref{tab:datascale} leads us to several conclusions: 1) The accuracy of person ReID through our self-supervised learning algorithm increases with the scale of pre-training data, implying that our method of learning self-supervised features is more suitable for large-scale unlabeled data. Furthermore, the larger scale of the data, the higher the Reid accuracy; 2) When the data scale is relatively small, initializing with ImageNet pre-training parameters offers significant improvement; 3) Selecting the top data according to the CFS ranking in TransReID-SSL \cite{reid_ssl} outperforms random selection, indicating a certain correlation between CFS and final accuracy. However, this selection method did not achieve the result stated in the TransReID-SSL paper, where CFS (50\%) surpassed the accuracy of the full dataset.

\begin{table}[h!t]
\centering
\caption{\label{tab:datascale}Ablation studies about the amount of pre-trainingdatasets.} 
\begin{tabular}{l|l|l|ll} 
\toprule
Backbone     & w/ PT. & Scale      & mAP  & Rank-1 \\ 
\midrule
ViT-S/16 & \XSolidBrush       & 3\%      & 54.2 & 75.7   \\
ViT-S/16 & \XSolidBrush       & 10\%     & 65.8 & 84.0   \\
ViT-S/16 & \XSolidBrush       & 25\%     & 66.1 & 83.9   \\
ViT-S/16 & \XSolidBrush       & 50\%     & 70.5 & 87.1   \\
\rowcolor{TableColor}
ViT-S/16 & \XSolidBrush       & 100\%    & 74.3 & 89.2   \\ \hline
ViT-S/16 & \Checkmark       & 10\%     & 66.7 & 85.5       \\ \hline
ViT-S/16 & \Checkmark       & 10\%+CFS \cite{reid_ssl} & 67.5 & 84.9       \\ \hline
ViT-S/16 & \Checkmark       & 50\%+CFS \cite{reid_ssl} & 73.8 & 88.6  \\ 
\bottomrule
\end{tabular}
\end{table}

\subsection{Visualization Analysis}
\begin{figure}[htp]
\centering
\includegraphics[width=0.8\linewidth]{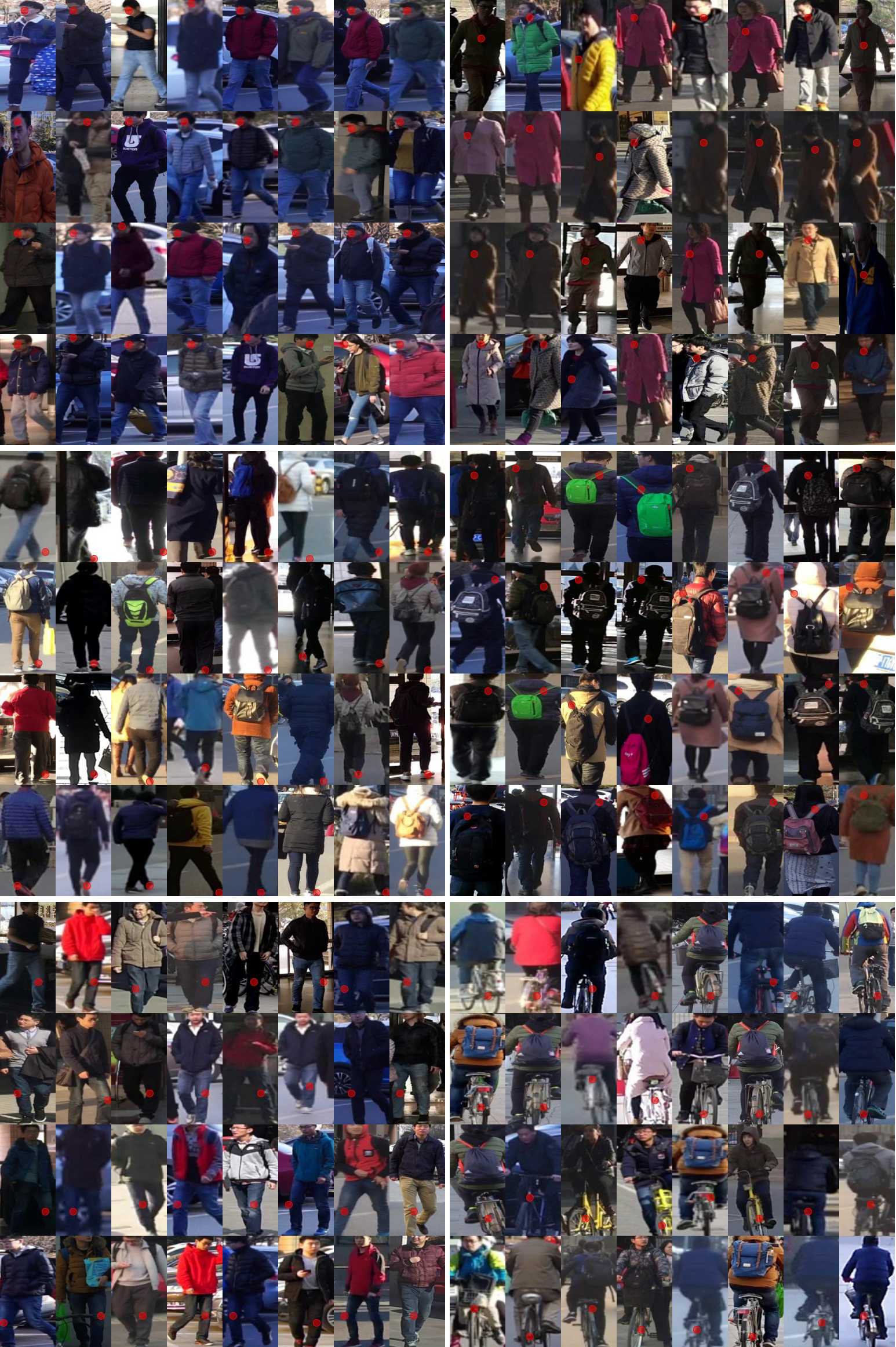}
\caption{\label{fig:vis_patch}Visualization for pattern layout of patch tokens cluster.}
\end{figure}

To better understand the human features learned by the pre-trained model, we conducted a visualization analysis of the pre-trained ViT-S/16 model on the MSMT17 \cite{MSMT17} dataset. This was pre-trained based on LUPerson, where we analyzed pattern layout for the patch tokens, visualized self-attention maps, and explored feature correlations.
\paragraph{Pattern Layout for Patch Tokens.} Fig.~\ref{fig:vis_patch} offers a visual representation of the unsupervised clustering results for the output features $y^{[patchs]}$ from PersonViT. Each subgraph represents a cluster, with the red dots marking the position of the image patches that belong to the cluster. The left half of the figure portrays that the features $y^{[patchs]}$ extracted by the PersonViT model can cluster key human body parts such as faces, feet, and knee joints. The right half of the figure demonstrates that the features $y^{[patchs]}$ can cluster the neck of human body, backpacks, and their straps. These results substantiate that the PersonViT pre-training model can autonomously extract fine-grained features of crucial human body parts and nearby ancillary objects of the human body. This automatic extraction capability hugely contributes to improving the accuracy of person ReID. For instance, automated positional awareness of the neck could enhance the discriminability of personal ornaments like necklaces; auto-location of face and head can increase the recognition of hairstyles; auto-detection of the feet can significantly improve the recognizability of shoes.

\begin{figure}[htp]
\centering
\includegraphics[width=0.8\linewidth]{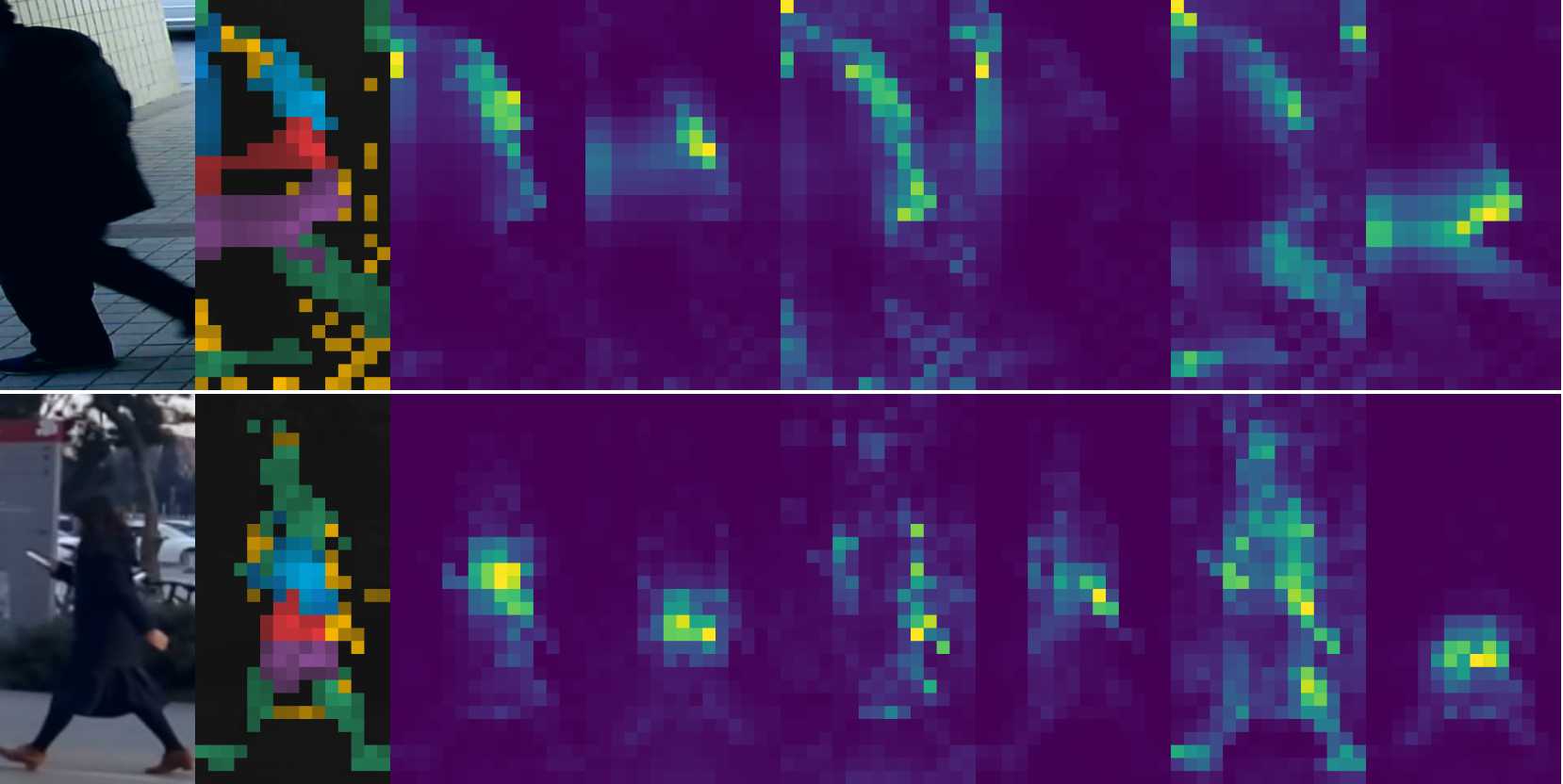}
\caption{\label{fig:vis_attention}Visualization for self-attention map from complex background.}
\end{figure}

\paragraph{Self-attention Visualizations.} As depicted by Fig.~\ref{fig:vis_attention}, the self-attention view confirms that the pre-training model can effectively extract body contours, even in intricate situations such as occlusions, fragmentations, or misalignments of human body, by dismissing the impact of irrelevant backgrounds. This ability explains the significant improvement of our algorithm on the Occluded-Duke dataset \cite{miao2019pose}.

\begin{figure}
\centering
\includegraphics[width=0.8\linewidth]{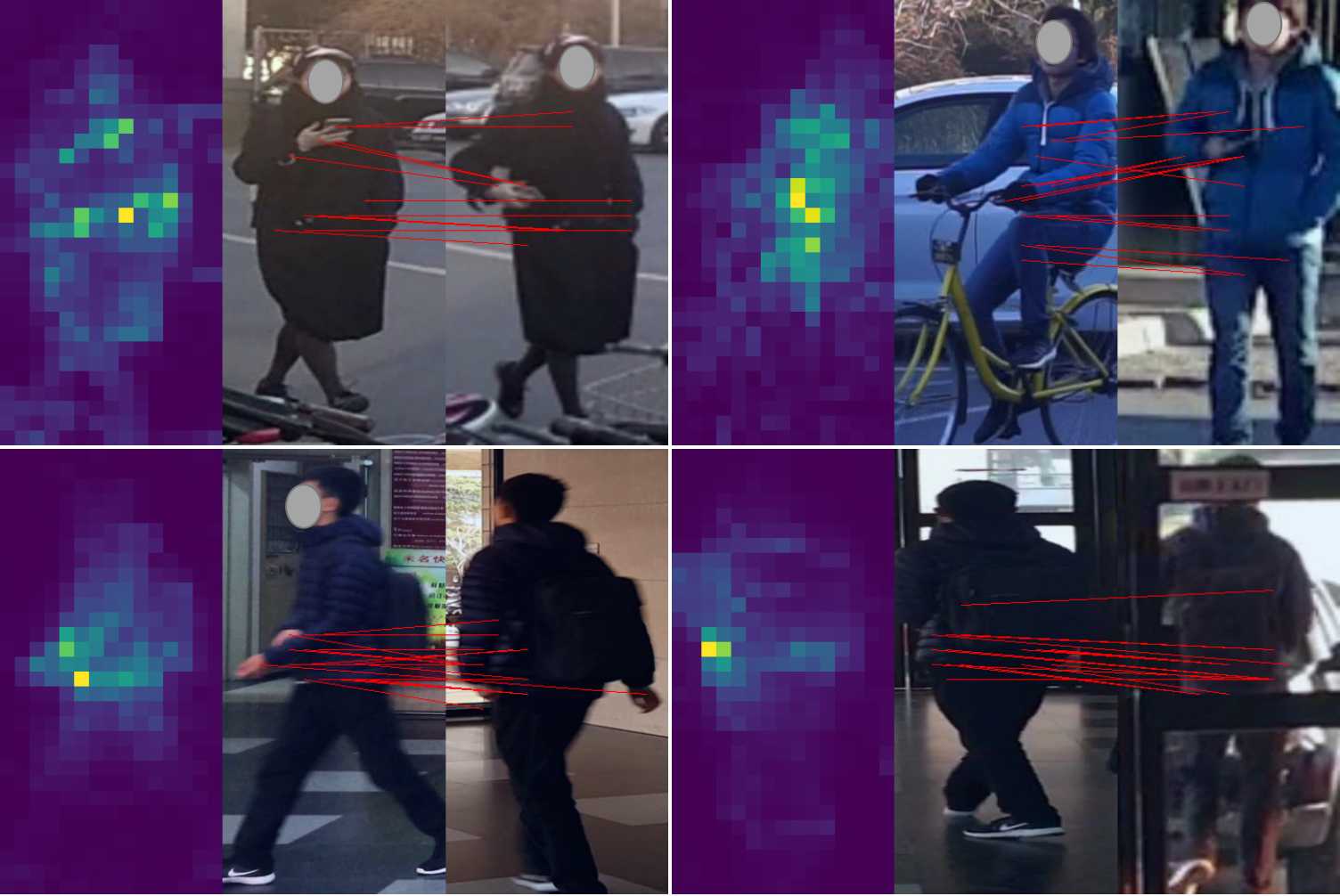}
\caption{\label{fig:vis_corresp}Visualization for sparse correspondence between two images of one person.}
\end{figure}

\paragraph{Feature Correlation Analysis.} The feature correlation assessment depicted in Fig.~\ref{fig:vis_corresp} indicates that our pre-training model can proficiently capture the relationships between features across various images of the same identity, even under significant deformations like turning around or cycling.


\section{Conclusion}

In this paper, we proposed a large-scale self-supervised person pre-training method, PersonViT, which introduces masked image modeling on the basis of contrastive learning. Through pre-training on a large-scale unlabeled human image dataset, PersonViT can effectively extract rich, highly discriminative, local fine-grained human features, and has achieved a significant accuracy improvement in person ReID tasks. Experiments show that even when the backbone network adopts a smaller vanilla ViT-S/16 model, the final recognition accuracy can further improve as the size of the pre-training dataset increases. Given that the cost of obtaining a large amount of unlabeled human pre-training data in real scenes is relatively low, this method has the potential for widespread application in practice, enhancing the effectiveness of re-identification algorithms in different scenarios.

However, like other self-supervised pre-training algorithms, PersonViT also faces the problem of high pre-training computational overhead, making the pre-training cycle longer in cases of limited computational resources. Although the frequency of updating the pre-training model for person ReID algorithms is lower than during supervised training, pre-training efficiency is still a key issue when PersonViT is extensively applied to vast amounts of unlabeled pre-training data from real-life situations. This problem can be addressed in several ways: 1) Choosing a more lightweight ViT model as the backbone network; 2) Adopting a masking method similar to MAE \cite{he2022mae}, where masked tokens are discarded (MAE experiments demonstrate that 75\% of tokens can be abandoned) and do not participate in the pre-training of the backbone network, which can improve pre-training efficiency; 3) Researching more efficient pre-training data filtration methods to reduce the size of the pre-training dataset without lowering accuracy, thus increasing pre-training efficiency; and 4) Incremental pre-training related research, meaning keeping the original features of the pre-training model and conducting online incremental pre-training learning for the additional pre-training data to update and evolve the pre-training model.

\newpage
\bibliographystyle{ieeetr}
\bibliography{nips.bib}









\end{document}